\documentclass[10pt,twocolumn,letterpaper]{article}

\usepackage[pagenumbers]{wacv} 
\usepackage[most]{tcolorbox} 
\usepackage{enumitem}
\usepackage{tikz}
\usepackage{graphicx}
\usepackage{subcaption}
\usepackage[ruled,vlined]{algorithm2e}

\usepackage{booktabs}
\usepackage{tabularx}
\usepackage{makecell}
\usepackage{ragged2e} 

\usepackage{bm}
\usepackage{float}
\usetikzlibrary{shapes.geometric, arrows, positioning, calc} 
\definecolor{wacvblue}{rgb}{0.21,0.49,0.74}
\usepackage[pagebackref,breaklinks,colorlinks,allcolors=wacvblue]{hyperref}

\def\wacvPaperID{3108} 
\def\confName{WACV}
\def\confYear{2026}

\title{BoardVision: Deployment-ready and Robust Motherboard Defect Detection with YOLO+Faster-RCNN Ensemble}

\author{
Brandon Hill\\
University of Maryland, Baltimore County\\
{\tt\small bhill@umbc.edu}
\and
KMA Solaiman\thanks{Corresponding Author}\\
University of Maryland, Baltimore County\\
{\tt\small ksolaima@umbc.edu}\\
}

\begin{document}
\maketitle
\begin{abstract}
Motherboard defect detection is critical for ensuring reliability in high-volume electronics manufacturing. While prior research in PCB inspection has largely targeted bare-board or trace-level defects, assembly-level inspection of full motherboards inspection remains underexplored. In this work, we present \textbf{BoardVision}, a reproducible framework for detecting assembly-level defects such as missing screws, loose fan wiring, and surface scratches. We benchmark two representative detectors - YOLOv7 and Faster R-CNN, under controlled conditions on the MiracleFactory motherboard dataset, providing the first systematic comparison in this domain. To mitigate the limitations of single models, where YOLO excels in precision but underperforms in recall and Faster R-CNN shows the reverse, we propose a lightweight ensemble, \textbf{Confidence-Temporal Voting }(CTV Voter), that balances precision and recall through interpretable rules. We further evaluate robustness under realistic perturbations including sharpness, brightness, and orientation changes, highlighting stability challenges often overlooked in motherboard defect detection. Finally, we release a deployable GUI-driven inspection tool that bridges research evaluation with operator usability. Together, these contributions demonstrate how computer vision techniques can transition from benchmark results to practical quality assurance for assembly-level motherboard manufacturing.
\end{abstract}
    

\section{Introduction}

Motherboards are the backbone of modern electronics, and their reliability is essential for large-scale manufacturing. Even subtle assembly-level defects such as missing screws, loose wiring, or scratches, can compromise system reliability and incur significant costs. Automated optical inspection (AOI) has long been used to detect surface-level anomalies, but most prior research and industrial solutions have concentrated on bare-board or trace-level PCB inspection \cite{anitha2017survey,ma2017defect,lv2024dataset}, leaving motherboard inspection comparatively underexplored.

Deep learning has advanced vision-based inspection with detectors such as Faster R-CNN \cite{ren2015faster} and YOLO variants \cite{wang2022yolov7}. Extensions of YOLOv7 \cite{yang2023yolov7,yang2024improved,dehaerne2023yolov7opt,satsangee2024defect}, YOLOv5-based models \cite{shen2024defect,tang2023pcb}, and SSD variants \cite{liu2022ssdpcb} have shown strong results for PCB and related defects.
But most studies end with accuracy reports on curated datasets, with limited focus on robustness to factory perturbations like lighting, blur, or orientation \cite{Park2023Sensors}, and few translate into deployable operator-facing systems.

Single detectors also exhibit trade-offs: higher precision or higher recall, but rarely both are achieved together under imperfect dataset distributions. Leveraging multiple models together in a lightweight voting ensemble could provide a more balanced trade-off, motivating our design.

To address these gaps, we introduce \textbf{BoardVision}, a reproducible motherboard defect detection framework designed with both research benchmarking and industrial deployment in mind. Our contributions are four-fold:  
\begin{enumerate}
    \item the first controlled benchmarking of YOLOv7 and Faster R-CNN on an assembly-level motherboard dataset (MiracleFactory) \cite{xin2023dataset},  
    \item the first robustness-oriented ensemble in this space, a lightweight Voter model with interpretable rules that balances precision and recall across detectors under diverse perturbations,  
    \item comprehensive robustness evaluation under sharpness, brightness, and orientation perturbations, and  
    \item the first deployable GUI-driven inspection system for motherboard defects, capable of both live and offline inference.
\end{enumerate}

By uniting rigorous benchmarking, robustness evaluation, and practical deployment, BoardVision demonstrates how vision research can transition from isolated metrics to meaningful, real-world quality assurance.

\section{Related Work}
 
Automated inspection of bare-board and trace-level printed circuit boards (PCBs) has progressed from rule-based techniques such as template matching and pixel differencing, which were fast but brittle to lighting and alignment variations \cite{anitha2017survey,ma2017defect}, to classical machine learning approaches that relied on handcrafted descriptors such as HOG and SIFT with classifiers like SVM or k-NN \cite{iwahori2018sift,kaya2022detection}. While these pipelines achieved moderate accuracy, they struggled particularly with subtle or small defect classes and were quickly surpassed by deep CNN-based methods \cite{kaya2022detection}.

Deep learning has since become the dominant paradigm, with convolutional detectors applied across public PCB datasets such as DeepPCB \cite{deeppcb}, DsPCBSD+ \cite{lv2024, lv2024dataset}, and HRIPCB \cite{hripcb_dataset}. While these benchmarks enabled progress, they primarily emphasize trace-level or component-level anomalies. In contrast, assembly-level defects in motherboards (e.g., missing screws, detached fan ports, scratches) remain underexplored \cite{xin2023dataset, shen2024defect}, motivating use of the MiracleFactory dataset \cite{xin2023dataset}.  

Among detection families, two-stage models like Faster R-CNN \cite{ren2015faster} have offered strong accuracy but suffer from slower inference and reduced stability under class imbalance. In contrast, one-stage models such as the YOLO family has shown a strong real-time detector performance across vision benchmarks \cite{wang2022yolov7}. Recent studies have shown that YOLOv7 achieves state-of-the-art accuracy and throughput across PCB and related manufacturing tasks: \cite{yang2023yolov7} improved YOLOv7 with attention modules for fine-grained PCB defect detection, \cite{satsangee2024defect} applied YOLOv7 to automated remanufacturing defects, and  \cite{dehaerne2023yolov7opt} optimized YOLOv7 for semiconductor wafer inspection. 
However, few studies directly compare YOLOv7 and Faster R-CNN under matched setups for motherboard-level defects.

More recent studies address broader inspection challenges. \cite{Li2021TCPMT} proposed an adaptive YOLOv2+Faster R-CNN ensemble for DIP soldering, highlighting accuracy degradation under lighting and orientation variations and addressing it with continual self-adaptation. \cite{Park2023Sensors} further analyzed how lighting, contamination, and pose shifts impact defect detection models, emphasizing the need for robustness-oriented evaluation. Ensemble approaches such as voting or hybrid pipelines have also been explored \cite{law2024ensemble}, though practical deployment remains limited.
 
Finally, most PCB defect detection studies conclude with benchmark metrics, without advancing to deployable tools or operator-facing systems. In contrast, our work provides both rigorous benchmarking and practical utility: we deliver the first head-to-head evaluation of YOLOv7 and Faster R-CNN on an assembly-level motherboard dataset, introduce a lightweight ensemble (CTV Voter) with interpretable rules to enhance robustness, and release a GUI-driven inspection tool. This combination bridges academic evaluation with factory-floor usability, positioning BoardVision as both a reproducible research artifact and a practical quality assurance aid. 

\section{System Overview}

BoardVision is designed as a modular, deployment-ready pipeline for automated motherboard defect detection. The full system is shown in Fig.~\ref{fig:system_overview}, which integrates three key stages: input processing, ensemble inference, and operator-facing visualization.

\textbf{Input and Preprocessing.} Images or video streams (either offline files or live camera feeds) are first passed through a preprocessing stage where YOLOv7 and Faster R-CNN detectors are initialized. These models then run on a CPU or GPU to produce bounding boxes and their corresponding confidence scores.

\textbf{Ensemble Inference.} Outputs from the detectors are reconciled by the Confidence–Temporal Voting (CTV) module. This layer matches detections across models using IoU criteria and applies either solo rules (e.g., high-confidence overrides, model preference) or agreement fusion (confidence- and F1-weighted averaging) before applying non-max suppression. The details of CTV are described in Section~\ref{sec:voteralgorithm}.

\textbf{Visualization and GUI.} The final predictions are overlaid on the input stream and presented through a PySide6-based GUI. 
The system supports two execution modes: 
(1) \emph{streaming mode}, for live video inspection, and 
(2) \emph{file mode}, for batch evaluation of offline images and videos.
The interface logs model disagreements, exposes ensemble decisions, and allows operators to adjust key parameters in real time.
Together, these components frame BoardVision as a practical QA tool, bridging algorithmic advances with user-facing deployment.

\section{Method: Voter Algorithm}  
\label{sec:voteralgorithm}

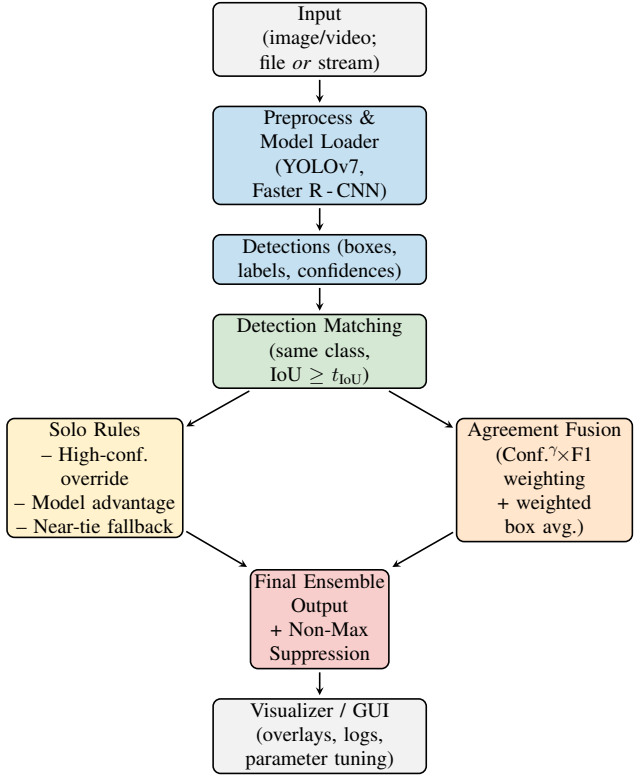
\begin{figure}[t]
\centering
\resizebox{\columnwidth}{!}{%
\begin{tikzpicture}[node distance=0.3cm]

\tikzstyle{block} = [rectangle, rounded corners, draw=black,
    text width=11.0em, text centered, minimum height=2em,
    font=\fontsize{11pt}{13pt}\selectfont, inner sep=2pt, outer sep=1pt, line width=0.3mm]
\tikzstyle{arrow} = [thick,->,>=stealth, line width=0.3mm, shorten <=1pt, shorten >=1pt]

\definecolor{boxblue}{HTML}{C5E0F1}
\definecolor{boxgreen}{HTML}{D5E8D4}
\definecolor{boxyellow}{HTML}{FFF2CC}
\definecolor{boxorange}{HTML}{FFE6CC}
\definecolor{boxred}{HTML}{F8CECC}
\definecolor{boxio}{HTML}{F2F2F2}

\node (input) [block, fill=boxio] {Input\\(image/video; file \textit{or} stream)};
\node (loader) [block, fill=boxblue, below=0.5cm of input] {Preprocess \& Model Loader\\(YOLOv7, Faster R\,-\,CNN)};

\node (detections) [block, fill=boxblue, below=0.5cm of loader] {Detections (boxes, labels, confidences)};
\node (matching)   [block, fill=boxgreen, below=0.5cm of detections] {Detection Matching\\(same class, IoU $\ge$ $t_{\text{IoU}}$)};

\node (solo)  [block, text width=9.0em, fill=boxyellow, below left=0.5cm and 0.5cm of matching]
              {Solo Rules\\-- High-conf. override\\-- Model advantage\\-- Near-tie fallback};

\node (fusion)[block, text width=9.0em, fill=boxorange, below right=0.5cm and 0.5cm of matching]
              {Agreement Fusion\\(Conf.$^\gamma\!\times$F1 weighting\\+ weighted box avg.)};

\coordinate (mid_solo_fusion) at ($ (solo.south)!0.5!(fusion.south) $);
\node (final) [block, fill=boxred, text width=7.0em, below=0.5cm of mid_solo_fusion]
              {Final Ensemble Output\\+ Non-Max Suppression};

\node (gui) [block, fill=boxio, below=0.5cm of final]
            {Visualizer / GUI\\(overlays, logs, parameter tuning)};

\draw [arrow] (input) -- (loader);
\draw [arrow] (loader) -- (detections);
\draw [arrow] (detections) -- (matching);
\draw [arrow] (matching) -- (solo);
\draw [arrow] (matching) -- (fusion);
\draw [arrow] (solo) -- (final.north west);
\draw [arrow] (fusion) -- (final.north east);
\draw [arrow] (final) -- (gui);

\end{tikzpicture}%
}
\caption{BoardVision pipeline: YOLOv7 and Faster R-CNN outputs are reconciled by the Confidence–Temporal Voting ensemble and visualized through a GUI for operator-facing deployment.}
\label{fig:system_overview}
\end{figure}

While YOLOv7 achieves high overall accuracy and Faster R-CNN produces more 
stable detections under class imbalance, each exhibits complementary failure 
modes. To exploit these strengths, we introduce a lightweight ensemble strategy 
called \textbf{Confidence-Temporal Voting (CTV)}. Fig.~\ref{fig:system_overview} includes the CTV workflow. The full pseudocode is provided in Algorithm~\ref{algo:ctv} in Appendix.

\subsection{Parameter Intuition} 
CTV is governed by a small set of tunable and interpretable parameters that control 
pairing, confidence weighting, and solo-detection rules:

\begin{itemize}
  \item \textbf{IoU threshold} ($t_{IoU}$): minimum overlap required to pair detections across models.
  \item \textbf{Confidence exponent} ($\gamma$): emphasizes high-confidence predictions.
  \item \textbf{Class F1 margin} ($f1\_margin$): tolerance for treating two models as equivalent on a class.
  \item \textbf{Solo confidence thresholds}: govern when to retain unmatched detections - $conf\_thresh, solo\_strong$.
\end{itemize}


\subsection{Detection Matching} 
Per-frame detections from YOLOv7 and Faster R-CNN are represented as bounding 
boxes with class labels and confidence scores. 
Let i and j be the indices for the detections from the YOLOv7 and Faster R-CNN models, respectively.
For each YOLO detection $y_i$, we attempt to pair it with a Faster R-CNN 
detection $f_j$ of the same class if their intersection-over-union (IoU) 
exceeds a threshold ($t_{IoU}$). 
Two cases then follow before the final decision: 

\begin{itemize}
  \item \textbf{Agreement:} 
If a valid agreement pair $(y_i, f_j)$ is found, the detections are merged 
through a weighted voting scheme that balances instance confidence with 
class-level reliability.
A worked example of this fusion process is provided in Appendix A.

  \item \textbf{Solo:} If no valid pair is found, we apply a solo-labeling heuristic 
  that decides whether to retain the detection based on interpretable rules.
\end{itemize}

\subsubsection{Agreement Cases} 

\paragraph{Confidence-weighted scoring}
    An exponent $\gamma$ is applied to model confidences, amplifying the influence of higher-confidence predictions during fusion.
For an agreement pair of class $c$, we assign each model a \textbf{fusion score}:
    \begin{align}
    S_{\text{YOLO}}(i) &= \big(p^{\text{YOLO}}_i\big)^\gamma \cdot F1_{\text{YOLO},c} \\
    S_{\text{FRCNN}}(j) &= \big(p^{\text{FRCNN}}_j\big)^\gamma \cdot F1_{\text{FRCNN},c}
    \end{align}
    
    where
        $p^\text{YOLO}_i, p^\text{FRCNN}_j$ are the model confidence scores for YOLO and FRCNN detections, and 
        $F1_{\cdot,c}$ is the model’s validation F1 score for class $c$.
    
    This formula balances instance-level confidence (per detection) with class-level trustworthiness (per model). A model that performs better historically on class $c$ exerts more influence, even if its raw confidence is slightly lower.

\paragraph{Bounding Box for Fused Prediction}
The final bounding box is a weighted average between $S_{\text{YOLO}}$ and $S_{\text{FRCNN}}$, weighted by bounding boxes from YOLO ($b_Y$) and FRCNN ($b_R$):
\begin{align}
b^\star = \frac{S_{YOLO}\cdot b_Y + S_{FRCNN}\cdot b_R}{S_{YOLO}+S_{FRCNN}}
\end{align}
Finally, the fused confidence is calculated as $\max(p_Y,p_R)$.\\

\subsubsection{Solo Cases} 
When no valid pair is found, CTV applies three interpretable rules to decide  whether to keep a detection:
\begin{enumerate}[label=(\Roman*)]
  \item \textbf{High-confidence override:} Retain any detection with confidence above 
        the \texttt{solo\_strong} threshold.
  \item \textbf{Model advantage:} Retain if the model’s per-class F1 score 
        is higher than its competitor and the confidence exceeds \texttt{conf\_thresh}.
  \item \textbf{Near-tie fallback:} Retain if the two models perform similarly on a specific 
        class, in terms of their F1-score (within $f1\_margin$) and the confidence is at least 0.95.
\end{enumerate}


\section{Experimental Setup and Results} %
\label{sec:results}

\subsection{Dataset} 
We evaluate on the publicly available \textbf{MiracleFactory Motherboard Defect Dataset} \cite{xin2023dataset}, 
which contains 389 high-resolution images and 2,860 annotated instances across 11 defect 
categories. This dataset was designed for visual quality assurance (QA) in motherboard manufacturing, but it 
poses several challenges: (1) the number of samples is modest compared to common detection 
benchmarks; (2) class distribution is highly imbalanced, with common classes such as 
\textit{Screws} dominating, while rare but operationally important categories such as 
\textit{Loose Screws} and \textit{CPU\_fan\_port\_detached} contain fewer than 100 
instances; and (3) many labels represent fine-grained, visually similar categories (e.g., different screw 
conditions), making accurate discrimination difficult under imbalance. Table~\ref{tab:defect_counts} summarizes the 
distribution.

\begin{table}[h]
\centering
\caption{Distribution of annotated instances across defect classes.}
\begin{tabular}{|l|c|}
\hline
\textbf{Class Name} & \textbf{Instance Count} \\
\hline
Screws & 806 \\
CPU\_FAN\_Screws & 685 \\
CPU\_FAN\_NO\_Screws & 326 \\
CPU\_fan & 313 \\
No\_Screws & 196 \\
CPU\_fan\_port & 159 \\
CPU\_FAN\_Screw\_loose & 99 \\
Scratch & 95 \\
Incorrect\_Screws & 63 \\
CPU\_fan\_port\_detached & 60 \\
Loose\_Screws & 58 \\
\hline
\end{tabular}
\label{tab:defect_counts}
\end{table}


\subsection{Training and Evaluation Protocols} 
We compare YOLOv7 and Faster R-CNN under matched training setups to ensure fairness. 
Both were trained for 50 epochs with early stopping. Models were trained with stochastic gradient descent and comparable learning 
rates, using input size of 640×640. YOLOv7 followed its official PyTorch implementation 
with a base learning rate of 0.01, while Faster R-CNN used a ResNet-50 FPN backbone via 
torchvision with a base learning rate of 0.005. 
For the CTV approach, we used 
$t_{IoU}=0.4$, $\gamma=2$, and $f1\_margin=0.05$, with 
$p^\text{YOLO}_i = 0.6, p^\text{FRCNN}_j = 0.9$, $conf\_thresh=0.6$ and $solo\_strong=0.95$ as default values.

Inference-time robustness tests uses image augmentations, including
horizontal flipping, increased sharpness, and varied brightness. To address class imbalance, no resampling 
was applied, reflecting real world deployment conditions.

Training was performed on a local workstation (NVIDIA GTX 1080 GPU with 8GB VRAM, AMD 
3900x CPU, 32GB RAM). Models were implemented in PyTorch 2.0 with CUDA 11.7.
Checkpoints were saved every epoch, and the model with the highest validation performance was selected 
for final evaluation. This ensured consistent convergence monitoring while enabling 
reproducibility of results.

We evaluate all models on a held-out test set of 45 images ($640\times640$), 
containing 365 annotated instances across all 11 classes. Standard object detection metrics were used to evaluate model performance \cite{Everingham2010VOC, Lin2014COCO}:

\begin{itemize}
  \item \textbf{mAP@0.5}: Mean Average Precision at an Intersection over Union (IoU) threshold of 0.5, measuring bounding box localization and classification accuracy.
  \item \textbf{mAP@0.5:0.95}: A stricter metric averaging mAP across IoU thresholds from 0.5 to 0.95 (in 0.05 increments), capturing both coarse and fine localization quality.
  \item \textbf{Precision} (\textbf{P}): Ratio of correctly predicted defect instances to total predictions, reflecting false positive rates.
  \item \textbf{Recall} (\textbf{R}): Proportion of ground truth instances correctly identified, reflecting false negative rates.
  \item \textbf{F1-score} (\textbf{F1}): Harmonic mean of precision \& recall.
  \item \textbf{FPS (frames/sec)}: The number of images processed per second, indicating real-time viability. 
\end{itemize}

\label{sec:results}

\subsection{Full Quantitative Results} 

Table~\ref{tab:voter_results} compares YOLOv7, Faster R-CNN (FRCNN), and the proposed voter ensemble (CTV) across aggregate detection metrics. As expected, YOLOv7 delivers the highest overall performance, achieving superior mAP@0.5, precision, and F1-score, while also running over twice as fast as Faster R-CNN. Faster R-CNN lags across most metrics, reflecting its sensitivity to small, cluttered objects and the dataset’s class imbalance. The ensemble improves upon YOLOv7 by balancing precision and recall, yielding the highest F1-score (0.964) and top precision (0.967). This confirms that the voter algorithm enhances stability without sacrificing YOLOv7’s high recall, and provides an interpretable pathway for integrating complementary detectors. 

\begin{table}[htbp]
    \centering
    \caption{Aggregate performance comparison across models.}
    \label{tab:voter_results}
    \begin{tabular}{|l|c|c|c|}
    \hline
        \textbf{Metric} & 
        \textbf{YOLOv7} & 
        \shortstack{\textbf{FRCNN}} &
        \shortstack{\textbf{CTV}} \\
    \hline
    mAP@0.5 & 0.914 & 0.766 & \textbf{0.921} \\
    mAP@0.5:0.95 & \textbf{0.606} & 0.495 & 0.604 \\
    Precision & 0.964 & 0.953 & \textbf{0.967} \\
    Recall & 0.956 & 0.718 & \textbf{0.962} \\
    Mean F1-score & 0.960 & 0.819 & \textbf{0.964} \\
    FPS (frames/sec) & 22--25 & 8--10 & 8--10 \\
    \hline
    \end{tabular}
\end{table}


Table~\ref{tab:f1_scores} provides a per-class breakdown of F1-scores across the 11 defect categories. YOLOv7 achieves near-perfect performance on most of the classes, including frequent categories such as \textit{Screws} and \textit{CPU\_FAN\_Screws}, as well as rare but operationally critical types like \textit{Loose\_Screws}. Faster R-CNN, in contrast, collapses on rare or visually subtle categories (e.g., \textit{CPU\_FAN\_NO\_Screws}), reflecting its sensitivity to class imbalance. The voter ensemble maintains YOLOv7’s strong scores while selectively improving specific categories, most notably \textit{No\_Screws}, where it raises F1 from 0.926 to \textbf{0.963} by selectively retaining high-confidence Faster R-CNN detections. These results confirm that the ensemble preserves YOLOv7’s dominance while mitigating edge-case failures that are critical in deployment.


\begin{table}[htbp]
    \footnotesize
    \centering
    \caption{Per-class F1-scores across models.} 
    \label{tab:f1_scores}
    \begin{tabular}{|l|c|c|c|}
        \hline
        \textbf{Class} & 
        \textbf{YOLOv7} & 
        \shortstack{\textbf{FRCNN}} &
        \shortstack{\textbf{CTV}} \\
        \hline
        CPU\_FAN\_NO\_Screws & \textbf{0.907} & 0.000 & \textbf{0.907} \\
        CPU\_FAN\_Screw\_loose & \textbf{0.933} & 0.615 & \textbf{0.933} \\
        CPU\_FAN\_Screws & \textbf{1.000} & 0.959 & \textbf{1.000} \\
        CPU\_fan & 0.987 & \textbf{1.000} & \textbf{1.000} \\
        CPU\_fan\_port & \textbf{0.974} & 0.789 & \textbf{0.974} \\
        CPU\_fan\_port\_detached & \textbf{0.667} & 0.462 & \textbf{0.667} \\
        Incorrect\_Screws & \textbf{0.857} & 0.667 & \textbf{0.857} \\
        Loose\_Screws & \textbf{1.000} & 0.800 & \textbf{1.000} \\
        No\_Screws & 0.926 & 0.851 & \textbf{0.963} \\
        Scratch & \textbf{0.900} & \textbf{0.900} & \textbf{0.900} \\
        Screws & \textbf{0.987} & 0.895 & \textbf{0.987} \\
        \hline
    \end{tabular}
\end{table}

Overall, these results show that 
ensemble recovers stability on difficult classes without hurting throughput. This validates the value of the proposed confidence-temporal voting scheme for industrial QA.

\paragraph{Failure Modes and Error Analysis.}
The per-class F1 scores (Table~\ref{tab:f1_scores}) and error profiles (Table~\ref{tab:error_profile}) highlight systematic strengths and weaknesses of each detector. YOLOv7 achieves strong separation on most categories, but struggles with \textit{CPU\_fan\_port\_detached}, where only 0.67 of samples are correctly identified, consistent with its lower F1 score for this class. Faster R-CNN shows broader distributed errors, with 103 false negatives overall, reflecting a strong recall deficit. In contrast, YOLOv7 produces more false positives (13) despite higher recall.

The proposed voter ensemble inherits YOLOv7’s precision while reducing Faster R-CNN’s false negatives, achieving 351 true positives with only 12 false negatives. Notably, categories such as \textit{No\_Screws} improve from 0.93 F1 with YOLOv7 to 0.96 with the ensemble, and \textit{Loose\_Screws} is preserved at 1.00 compared to 0.67 for Faster R-CNN. These results confirm that the ensemble provides more balanced error trade-offs, especially for rare but high-cost classes.
These stability gains extend under perturbations, as further analyzed in Section~\ref{sec:robustness}.

Full confusion matrices are included in the appendix for reference, but the main operational insights are captured by the compact FP/FN counts and per-class F1 scores.

\begin{table}[htbp] 
\centering
\caption{Error profile across models.}
\label{tab:error_profile}
\begin{tabular}{|l|l|c|c|c|}
\hline
\textbf{Method} & \textbf{TP} & \textbf{FP} & \textbf{FN} \\
\hline
YOLOv7
  & 349 & 13 & 16 \\
\hline
Faster R-CNN
  & 262 & 13  & 103 \\ 
\hline
Voter
  & 351 & 12 & 14 \\ 
\hline
\end{tabular}
\label{tab:error_profile}
\end{table}

\subsection{Qualitative Examples} 
\label{sec:qualitative}

\begin{figure}[h]
    \centering
    \begin{subfigure}[b]{0.9\linewidth}
        \includegraphics[width=\linewidth, trim=20 40 20 40, clip]{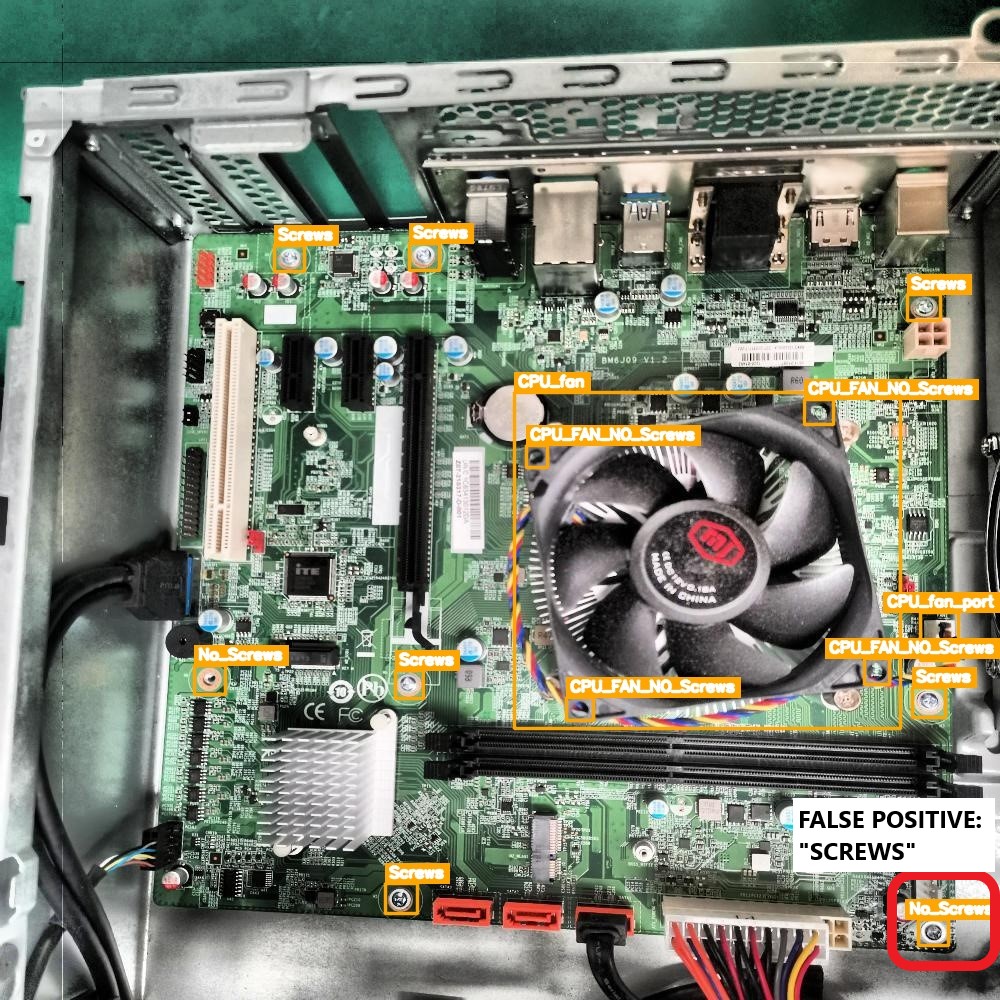}
        \caption{YOLOv7 – FP: Screws}
        \label{fig:yolo_qual_1}
    \end{subfigure}
    \hfill
    \begin{subfigure}[b]{0.9\linewidth}
        \includegraphics[width=\linewidth, trim=20 40 20 40, clip]{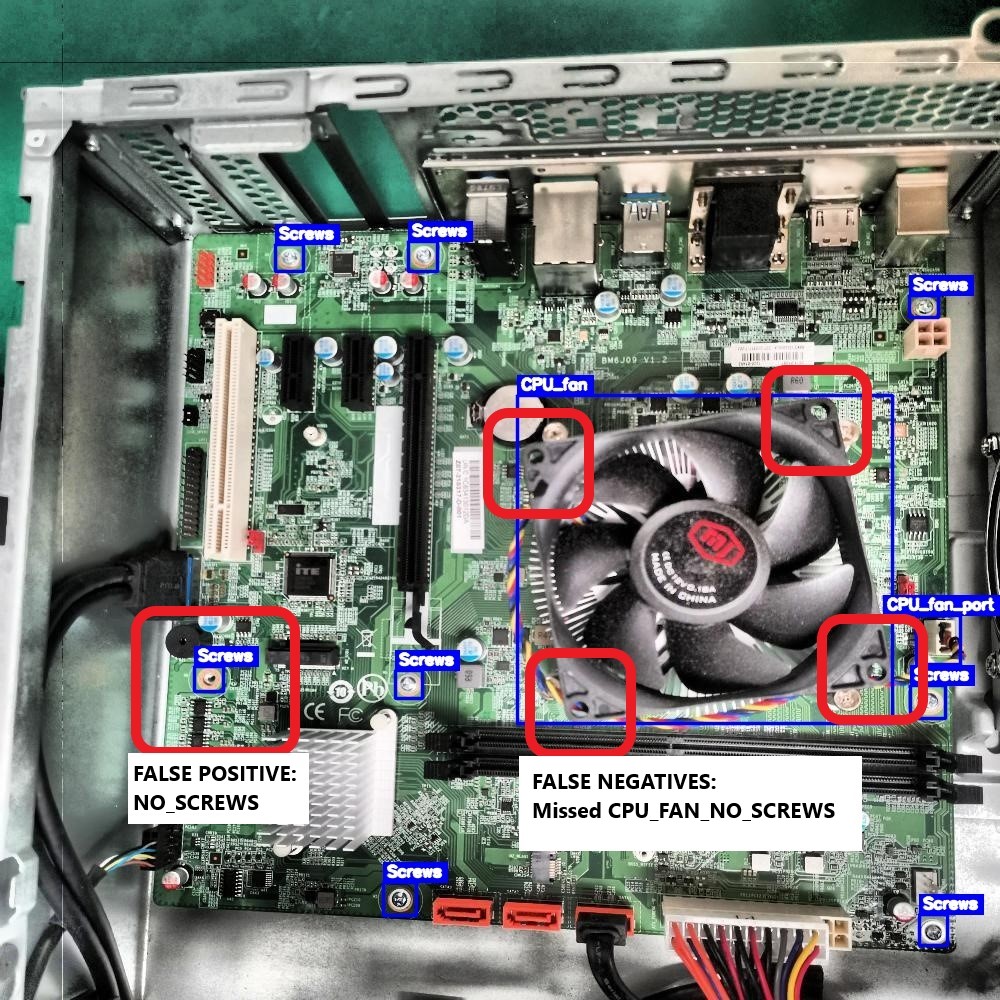}
        \caption{Faster R-CNN – FP: No Screws, FN: CPU\_FAN\_NO\_Screws}
        \label{fig:frcnn_qual_1}
    \end{subfigure}
    \caption{Annotated error cases for YOLOv7 and FRCNN. False positives (FP) and false negatives (FN) are annotated in red to illustrate specific failure modes.}
    \label{fig:false_case_examples}
\end{figure}

In this section we illustrate how each model performs on real motherboard images. 
By comparing side-by-side predictions from YOLOv7 and Faster R-CNN (Figure~\ref{fig:false_case_examples}) on the same test sample, we can better understand practical differences and failure modes in their behavior.

\begin{figure}[h]
  \centering
  \includegraphics[width=0.9\linewidth]{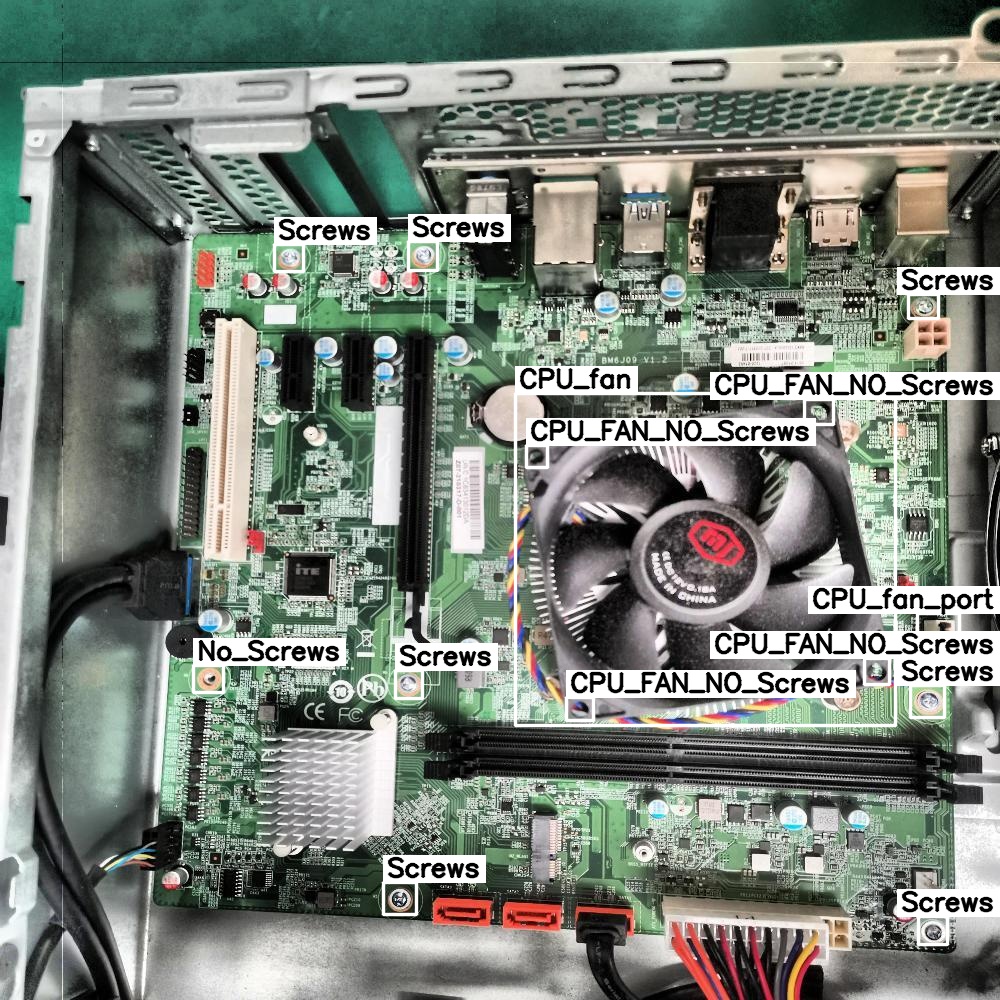}
  \caption{Ground truth annotations for the motherboard test image used in Figures~\ref{fig:false_case_examples} and \ref{fig:voter_qual_1}.} 
  \label{fig:ground_truth}
\end{figure}

\begin{figure}[h]
    \centering
    \includegraphics[width=0.9\linewidth]{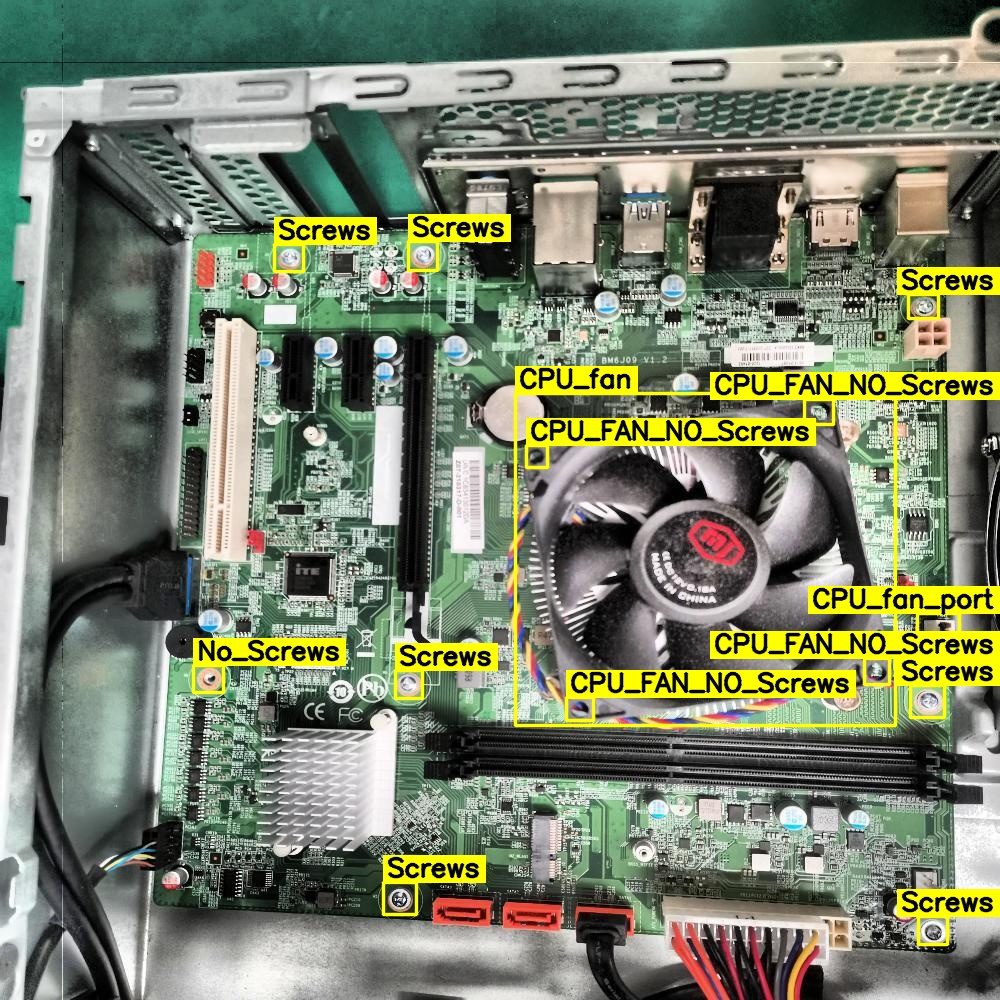}
    \caption{Voter Predictions for motherboard test image.}
    \label{fig:voter_qual_1}
\end{figure}

Both single models identify key components like \textit{Screws} and \textit{CPU\_fan}, but differences in granularity and class distinction are evident. In Figure~\ref{fig:yolo_qual_1}, YOLOv7 outputs precise, non-overlapping predictions with clear boundaries and minimal redundancy. It correctly identifies multiple \textit{Screws} regions and distinguishes \textit{CPU\_fan} components without overlap or redundancy. YOLOv7's predictions are visually coherent and class-consistent, reflecting its superior per-class F1 scores.

Figure~\ref{fig:false_case_examples} highlights typical failure modes. In both cases, a false positive for \textit{No\_Screws} appears in a region where screws exist, indicating that metallic reflections and shadows near screw holes can trigger confusion. Faster R-CNN also misses a true \textit{CPU\_FAN\_NO\_Screws} instance within the fan mount, suggesting reduced sensitivity to small absences in dense areas. YOLOv7 incorrectly labels an existing screw as \textit{No\_Screws} in a crowded region.

The voter ensemble consistently corrects these issues. By fusing detections from both models with agreement-aware scoring and cross-model suppression, the voter removes spurious \textit{No\_Screws} predictions, restores missed \textit{CPU\_FAN\_NO\_Screws} near the fan mount, and preserves confident \textit{Screws} and \textit{CPU\_fan} boxes. Qualitatively, Figure~\ref{fig:voter_qual_1} shows tighter boxes, fewer label flips, and better separation of visually similar classes.


\section{Robustness and Sensitivity Analysis}
\label{sec:sensitivity}

Beyond baseline performance, we analyze the stability of BoardVision under both external perturbations (image augmentations) and internal variations (ensemble parameters and ablation). 

\paragraph{Robustness Evaluation.}  
\label{sec:robustness}

To evaluate robustness, we tested all models on the original held-out test set and on augmented variants designed to simulate real-world perturbations: 
\begin{itemize}
    \item \textbf{flip}: Horizontal mirroring of images to simulate viewpoint changes and test invariance to orientation.  
    \item \textbf{increased sharpness}: via Gaussian unsharp masking to mimic sensor or compression artifacts.
    \item \textbf{increased brightness}: via linear-RGB exposure adjustment, simulating overexposed, well-lit conditions.  
    \item \textbf{decreased brightness}: using the same method, simulating underexposed or poorly lit conditions.  
\end{itemize}
Such perturbations are representative of deployment scenarios where camera streams may vary in lighting, sharpness, or orientation, particularly in low-cost or factory-floor settings. The voter was run with default parameters, with per-class weights initialized from preliminary F1 scores of the base models. Tables \ref{tab:yolo_robustness}-\ref{tab:voter_robustness} refer to N = Normal, F = Flip, SUp = Sharpness Up, BUp = Brightness Up, BDn = Brightness Down.

\begin{table}[htbp]
\footnotesize
\centering
\caption{YOLOv7 Performance under Perturbations.}
\label{tab:yolo_robustness}
\begin{tabular}{|l|c|c|c|c|c|c|}
\hline
\textbf{Metric} & \textbf{Mean ± Std} & \textbf{N} & \textbf{F} & \textbf{SUp} & \textbf{BUp} & \textbf{BDn} \\
\hline
P & 0.962 ± 0.006 & 0.964 & 0.969 & 0.950 & 0.964 & 0.967 \\
R    & 0.949 ± 0.009 & 0.956 & 0.953 & 0.934 & 0.951 & 0.959 \\
F1   & \textbf{0.958 ± 0.007} & 0.960 & 0.961 & 0.942 & 0.957 & 0.963 \\
\hline
FPS & \multicolumn{6}{c|}{22–25} \\
\hline
\end{tabular}
\end{table}

\begin{table}[htbp]
\footnotesize
\centering
\caption{Faster R-CNN Performance under Perturbations.} 
\label{tab:frcnn_robustness}
\begin{tabular}{|l|c|c|c|c|c|c|}
\hline
\textbf{Metric} & \textbf{Mean ± Std} & \textbf{N} & \textbf{F} & \textbf{SUp} & \textbf{BUp} & \textbf{BDn} \\
\hline
P & 0.954 ± 0.008 & 0.953 & 0.967 & 0.945 & 0.945 & 0.959 \\
R    & 0.713 ± 0.003 & 0.718 & 0.721 & 0.710 & 0.712 & 0.712 \\
F1   & 0.816 ± 0.005 & 0.819 & 0.826 & 0.811 & 0.812 & 0.818 \\
\hline
FPS & \multicolumn{6}{c|}{8-10} \\
\hline
\end{tabular}
\end{table}

\begin{table}[htbp]
\footnotesize
\centering
\caption{CTV Performance under Perturbations.}
\label{tab:voter_robustness}
\begin{tabular}{|l|c|c|c|c|c|c|}
\hline
\textbf{Metric} & \textbf{Mean ± Std} & \textbf{N} & \textbf{F} & \textbf{SUp} & \textbf{BUp} & \textbf{BDn} \\
\hline
P    & \textbf{0.964 ± 0.006} & 0.967 & 0.972 & 0.953 & 0.961 & 0.967 \\
R       & \textbf{0.954 ± 0.009} & 0.962 & 0.948 & 0.940 & 0.956 & 0.962 \\
F1      & \textit{0.957 ± 0.006} & 0.964 & 0.960 & 0.946 & 0.959 & 0.964 \\
\hline
FPS & \multicolumn{6}{c|}{8-10} \\ 
\hline
\end{tabular}
\end{table}

Across augmentations shown in Table \ref{tab:yolo_robustness}, YOLOv7 maintains strong Recall but suffers drops under \textit{Sharp\_Up} (F1 = 0.942), exposing some fragility to high-frequency distortions. 
Faster R-CNN degrades more substantially, with Mean F1 = 0.816. Its heavier reliance on region proposals makes it sensitive to small feature perturbations, leading to lower stability across all conditions. 
The Voter Ensemble in Table \ref{tab:voter_robustness} matches YOLOv7’s F1 (0.957 vs. 0.958) while delivering higher precision and lower variance. 
Across augmentations, the Voter’s F1 ranges from 0.946 (Sharp\_Up) to 0.964 (Normal), consistently narrowing YOLOv7’s fluctuations. Even when YOLOv7 dips under augmentation like in Sharp\_Up, the Voter stabilizes performance by leveraging cross-model agreement.
The voter’s stability across perturbations demonstrates robustness: even when a single model wavers, the ensemble preserves consistent performance, making it more reliable for deployment on noisy or imperfect inspection streams.

\paragraph{Sensitivity to Voter Parameters.} 
\begin{table}[ht]
\footnotesize
\centering
\caption{Sensitivity of CTV parameters across IoU, $\gamma$, and solo thresholds. Bold values indicate notable deviations.}
\label{tab:sensitivity}
\begin{tabular}{|lcccc|}
\hline
\shortstack{Voter Params} &
\shortstack{mAP@0.5} &
\shortstack{mAP@0.5:0.95} &
P &
R \\
\hline
\shortstack{$t_{IoU}$ = 0.3} & 0.921 & 0.604 & 0.967 & 0.962 \\
\shortstack{$t_{IoU}$ = 0.5} & 0.921 & 0.604 & 0.967 & 0.962 \\
\shortstack{$t_{IoU}$ = 0.7} & \textbf{0.919 $\downarrow$} & 0.604 & \textbf{0.949 $\downarrow$} & 0.962 \\
\hline
$\gamma$ = 0 & \textbf{0.917 $\downarrow$} & \textbf{0.605 $\uparrow$} & \textbf{0.964 $\downarrow$} & \textbf{0.959 $\downarrow$} \\
$\gamma$ = 1 & 0.921 & \textbf{0.605 $\uparrow$} & 0.967 & 0.962 \\
$\gamma$ = 2 & 0.921 & 0.604 & 0.967 & 0.962 \\
$\gamma$ = 3 & 0.921 & \textbf{0.596 $\downarrow$} & 0.967 & 0.962 \\
\hline
\textit{solo}=0.90 & \textbf{0.910 $\downarrow$} & \textbf{0.595 $\downarrow$} & \textbf{0.944 $\downarrow$} & 0.962 \\
\textit{solo}=0.95 & \textbf{0.912 $\downarrow$} & \textbf{0.597 $\downarrow$} & \textbf{0.956 $\downarrow$} & 0.962 \\
\textit{solo}=0.98 & 0.921 & 0.604 & 0.967 & 0.962 \\
\hline
\end{tabular}
\end{table}

As Table~\ref{tab:sensitivity} shows, CTV is stable across a wide range of parameters, 
but extremes degrade performance. IoU thresholds of 0.3–0.5 perform identically, while raising 
to 0.7 reduces precision (0.949) and mAP@0.5 (0.919). The exponent $\gamma$ performs best 
at 1–2, whereas $\gamma=0$ removes YOLO’s precision bias and lowers recall (0.959), and 
$\gamma=3$ suppresses mAP@0.5:0.95 (0.596). Solo thresholds below 0.98 admit too many weak 
detections, collapsing precision to 0.944–0.956. The default 
($t_{IoU}=0.5, \gamma=2, solo=0.98$) sits near the optimal balance.

\paragraph{Single-Factor Ablation of Ensemble Rules.} 

\begin{table}[htbp]
\footnotesize
\centering
\caption{Ablation study of Ensemble Components.}
\label{tab:ensemble-ablation}
\begin{tabular}{|p{0.25\linewidth}|c|c|c|c|}
\hline
\shortstack{Config} & 
\shortstack{mAP@0.5} &
\shortstack{mAP@0.5:0.95} &
\shortstack{P} & 
\shortstack{R} \\
\hline
\shortstack{Full Ensemble} & 0.921 & 0.604 & 0.967 & 0.962 \\ 
\hline
No High confidence & 0.910 & 0.595 & 0.944 & 0.962 \\
\hline
No Per-class F1 weighting   & 0.912 & 0.601 & 0.956 & 0.962 \\
\hline
\shortstack{Always-tie rule} & 0.912 & 0.597 & 0.956 & 0.962 \\ 
\hline
\end{tabular}
\end{table}

The ablation study in Table ~\ref{tab:ensemble-ablation} confirms that the full CTV ensemble achieves the strongest balance, with mAP@0.5 of 0.921 and precision/recall both above 0.96. Removing the high-confidence override notably degrades precision (-2.3\%) and reduces mAP, showing its importance for reducing false positives and driving measurable precision gains. Eliminating the per-class F1 weighting or relaxing tie rules causes only minor drops, indicating they fine-tune performance but are not as critical as the override. 
Overall, the study highlights that the ensemble rules improve robustness by preserving high-confidence detections and stabilizing recall without sacrificing precision.

\paragraph{Multi-Factor Stress Tests.} 
\begin{table}[ht]
\footnotesize
\centering
\caption{IoU and $\gamma$ sensitivity. Defaults: $t_{IoU}$=0.4, $\gamma=2$.}
\label{tab:ablation_iou_gamma}
\begin{tabular}{|c|c|c|c|c|c|}
\hline
\shortstack{${t_{IoU}}$} &
\shortstack{${\gamma}$} &
\shortstack{mAP@0.5} &
\shortstack{mAP@0.5:0.95} &
\shortstack{P} & 
\shortstack{R} \\
\hline
0.30 & 0 & \textbf{0.917 ↓} & 0.605 & 0.964 & \textbf{0.959 ↓} \\
0.30 & 1 & 0.921 & 0.605 & 0.967 & 0.962 \\
0.50 & 0 & \textbf{0.917 ↓} & 0.605 & 0.964 & \textbf{0.959 ↓} \\
0.50 & 1 & 0.921 & 0.605 & 0.967 & 0.962 \\
0.70 & 0 & 0.919 & \textbf{0.606 ↑} & \textbf{0.949 ↓} & 0.962 \\
0.70 & 3 & 0.919 & \textbf{0.596 ↓} & \textbf{0.949 ↓} & 0.962 \\
\hline
\end{tabular}
\end{table}

Table~\ref{tab:ablation_iou_gamma} shows that IoU and $\gamma$ interact smoothly near defaults, 
but extreme values combine poorly: e.g., $t_{IoU}$=0.70 with $\gamma=3$ reduces precision to \textbf{0.949 ↓} 
and mAP@0.5:0.95 to \textbf{0.596 ↓}.  

\begin{table}[ht]
\footnotesize
\centering
\caption{Impact of per-class F1 weighting under stress.} 
\label{tab:ablation_f1}
\begin{tabular}{|c|c|c|c|c|c|}
\hline
${t_{IoU}}$ & \textbf{${\gamma}$} & f1\_margin & mAP@0.5 & P & R \\
\hline
0.30 & 0 & Original & 0.917 & 0.964 & 0.959 \\
0.30 & 0 & All-ones & \textbf{0.764 ↓} & 0.970 & \textbf{0.718 ↓} \\
0.50 & 1 & Original & 0.921 & 0.967 & 0.962 \\
0.50 & 1 & All-ones & \textbf{0.768 ↓} & 0.974 & \textbf{0.721 ↓} \\
0.70 & 3 & Original & 0.919 & 0.949 & 0.962 \\
0.70 & 3 & All-ones & \textbf{0.767 ↓} & 0.970 & \textbf{0.718 ↓} \\
\hline
\end{tabular}
\end{table}

Table~\ref{tab:ablation_f1} reveals the hidden importance of per-class F1 weighting.  
When replaced with uniform all-ones weights, recall collapses by \textbf{24\% ↓} (0.962 $\to$ 0.718) and 
mAP@0.5 drops to \textbf{0.76 ↓}. This shows that F1 weighting is essential under class imbalance, 
ensuring rare but critical defects are not ignored.

The sensitivity and ablation analyses jointly demonstrate that CTV’s design is not redundant.  
The high-confidence override provides the main precision boost, while F1 weighting and tie rules 
act as stabilizers - modest in isolation but essential under stress. 

\section{BoardVision GUI Application}

To bridge benchmarking with deployment, we developed a lightweight graphical user interface (GUI) for BoardVision using Python’s PySide6/Qt framework. 

With a \textbf{3-column synchronized view}, the GUI enables side-by-side inspection of YOLOv7 and Faster R-CNN predictions, while transparently displaying the fused ensemble output in real time. This design operationalizes the performance trends described in Section~\ref{sec:results} and makes model disagreements directly observable to practitioners.

The interface supports flexible input sources (local files, webcams, network streams) and allows users to adjust key ensemble parameters such as \texttt{conf\_thresh}, \texttt{solo\_strong}, and \texttt{iou\_thresh}. User can also adjust runtime behavior (stride for frame skipping, pause/resume, or stop). Real-time logs and overlays expose the Confidence–Temporal Voting rules (Section~\ref{sec:voteralgorithm}), enabling practitioners to interpret how detections are promoted, suppressed, or fused during operation. \textbf{The left panel of the GUI} givess access to these controls.

\begin{figure}[htbp]
    \centering
    \includegraphics[width=1\linewidth]{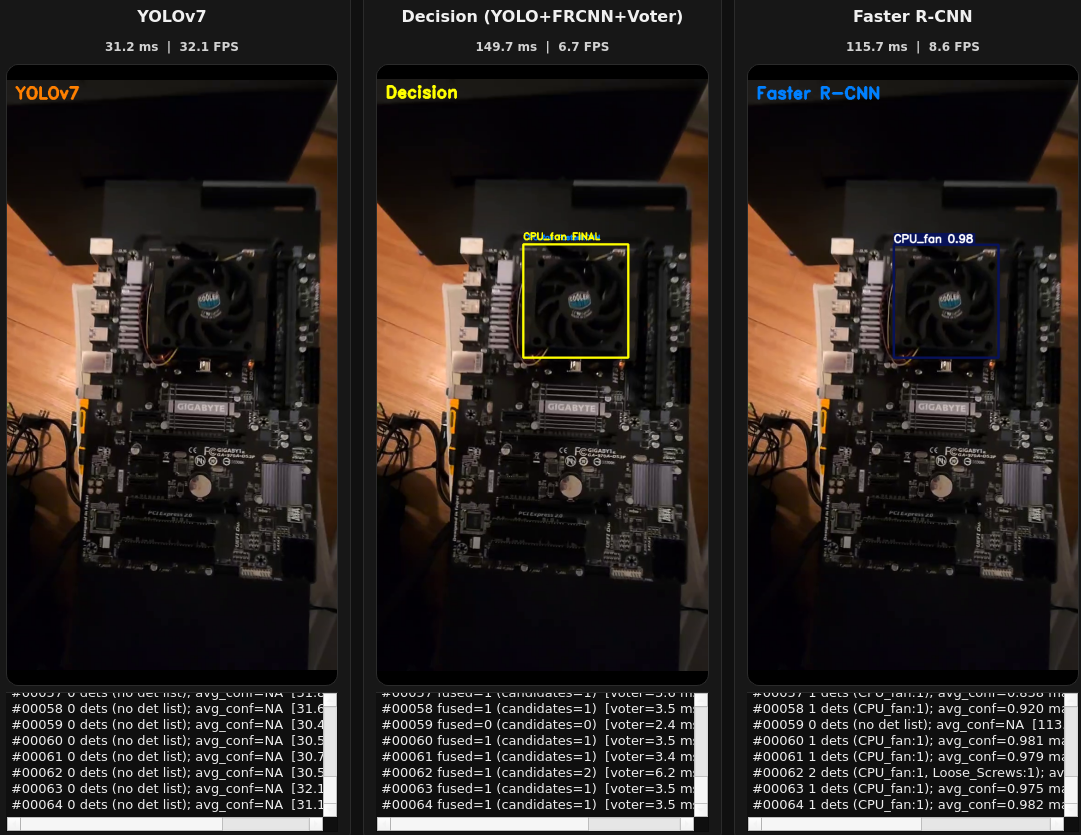}
    \caption{Live inference on a low-light, low-resolution stream. YOLOv7 misses a CPU fan, Faster R-CNN detects it with high confidence ($p=0.98$), and the voter ensemble fuses the result into the final decision (yellow box).}
    \label{fig:boardvision_inference_live}
\end{figure}


\section{Discussion and Future Work}
\label{sec:discussion}

Our study surfaced several practical lessons for applied defect detection. 
First, ensemble design can remain simple yet effective: interpretable rules such as high-confidence overrides and per-class weighting delivered stability without requiring complex meta-models. 
Second, dataset imbalance emerged as the primary bottleneck: rare but operationally critical defects consistently challenged single models, underscoring the need for either targeted data collection or ensemble mechanisms that preserve rare detections. 
Third, robustness analysis showed that stability across perturbations is as important as raw mAP, since factory conditions rarely match controlled benchmarks. 
Finally, deployment through the GUI demonstrated that transparency is crucial: visualizing candidate and fused boxes not only improved user trust but also made failure modes visible in ways that metrics alone cannot. 
These insights position BoardVision as a reproducible blueprint for building interpretable, deployment-ready vision systems in manufacturing.

Future work will focus on three directions: expanding dataset coverage to address class imbalance (particularly for rare but high-cost classes), adapting ensemble rules dynamically to improve generalization under distribution shifts, and integrating BoardVision more tightly into manufacturing pipelines through headless QA modules, automated reporting, and human-in-the-loop inspection.

\textbf{Ethics and Societal Impact.}
This work 
does not involve personal or sensitive data. Potential risks include over-reliance on automated inspection, which could miss rare but critical defects if deployed without human oversight. We recommend human-in-the-loop verification to ensure safety and reliability in practical use.

\textbf{Use of Large Language Models.} 
Parts of the BoardVision software were prototyped with assistance from OpenAI’s GPT-5 model. The LLM generated Python scaffolding for the graphical interface and the ensemble evaluation pipeline. All outputs were reviewed, debugged, and refactored by the authors to ensure correctness. Consistent with WACV policy, we take full responsibility for the final implementation.

\section{Conclusion}
\label{sec:conclusion}

This paper introduced BoardVision, an assembly-level  motherboard defect detection system that combines YOLOv7 and Faster R-CNN through a lightweight, interpretable ensemble. The proposed Confidence–Temporal Voting (CTV) framework reduced failure modes on rare but high-cost classes and delivered more stable performance than either detector alone. Experiments on the MiracleFactory dataset confirmed that the ensemble not only improved F1 and precision but also maintained robustness under perturbations. In live inference, the ensemble consistently outperformed either single model: running both detectors in parallel reduced indecisiveness on borderline cases, stabilized frame-to-frame decisions under low light or clutter, and filtered spurious boxes via cross-verification.

Beyond metrics, a deployment-oriented GUI demonstrated how the ensemble’s decision rules can be visualized and tuned in real time, bridging the gap between research benchmarks and practical motherboard inspection. We will publicly release the code, models, and the BoardVision software to support reproducibility in manufacturing QA and related applications. 
{
    \small
    \bibliographystyle{ieeenat_fullname}
    \bibliography{main}
}

\newpage
\appendix




\begin{algorithm}[h]
\caption{Confidence–Temporal Voting (CTV)} 
\KwIn{
YOLO detections $Y$, FRCNN detections $R$ (each as boxes $b$, label $\ell$, confidence $p$); \\
IoU threshold $t_{IoU}$; confidence exponent $\gamma$; \\
per-class validation scores $F1_{YOLO,c}$, $F1_{FRCNN,c}$; \\
solo thresholds $conf\_thresh$, $solo\_strong$; near-tie margin $f1\_margin$; \\
flag $fuse\_coords$ (average coordinates if true)
}
\KwOut{Final fused detections $F$}

$F \leftarrow \emptyset$ \;
Form candidate matches between $Y$ and $R$ with same label and $\mathrm{IoU}(b_Y,b_R)\ge t_{IoU}$ (greedy by IoU) \;

\ForEach{matched pair $(y_i, r_j)$}{
  $S_Y \leftarrow (p_Y)^{\gamma}\cdot F1_{YOLO,\ell}$,\quad $S_R \leftarrow (p_R)^{\gamma}\cdot F1_{FRCNN,\ell}$ \;
  \eIf{$fuse\_coords$}{
    $b^\star \leftarrow \dfrac{S_Y\, b_Y + S_R\, b_R}{S_Y + S_R}$\;
  }{
    $b^\star \leftarrow \arg\max_{b \in \{b_Y,b_R\}} \{S_Y,S_R\}$\;
  }
  $\ell^\star \leftarrow \ell$,\quad $p^\star \leftarrow \max(p_Y,p_R)$ \;
  Append $(b^\star,\ell^\star,p^\star)$ to $F$ \;
  Mark $y_i$ and $r_j$ as used \;
}

\ForEach{unmatched detection $d=(b,\ell,p)$ in $Y \cup R$}{
  let $m \in \{\text{YOLO},\text{FRCNN}\}$ be the model of $d$, and $m'$ the other model \;
  \If{$p \ge solo\_strong$}{Append $d$ to $F$; continue}
  \If{$F1_{m,\ell} > F1_{m',\ell}$ \textbf{and} $p \ge conf\_thresh$}{Append $d$ to $F$; continue}
  \If{$|F1_{\text{YOLO},\ell} - F1_{\text{FRCNN},\ell}| \le f1\_margin$ \textbf{and} $p \ge 0.95$}{Append $d$ to $F$; continue}
}

Apply class-wise Non-Maximum Suppression to $F$; return $F$ \;
\label{algo:ctv}
\end{algorithm}

\begin{figure}[h]
    \centering
    
\begin{tcolorbox}[title=Example of Fusion, colback=gray!5,colframe=gray!40!black]
YOLOv7 predicts [100,100,200,200] with confidence 0.9, Faster R-CNN predicts 
[110,105,195,205] with confidence 0.8. After weighting with class-level F1 and 
$\gamma=1.5$, YOLO’s score is 0.75, FRCNN’s 0.54.

    Weighting the coordinates yields:
    \[
    x_1^\star = \frac{0.75\cdot100+0.54\cdot110}{0.75+0.54}\approx104.2,\quad y_1^\star\approx102.0,
    \]
    and similarly for $(x_2,y_2)$. 
    So The fused box lies at 
[104.2, 102.0, ...], closer to YOLO’s prediction, reflecting its higher score, while still adjusting toward FRCNN’s box. 
\end{tcolorbox}
\caption{CTV Fusion Example.}
    \label{fig:ctv-fusion-example}
\end{figure}

\begin{figure*}[htbp]
    \centering
    \includegraphics[width=0.7\linewidth]{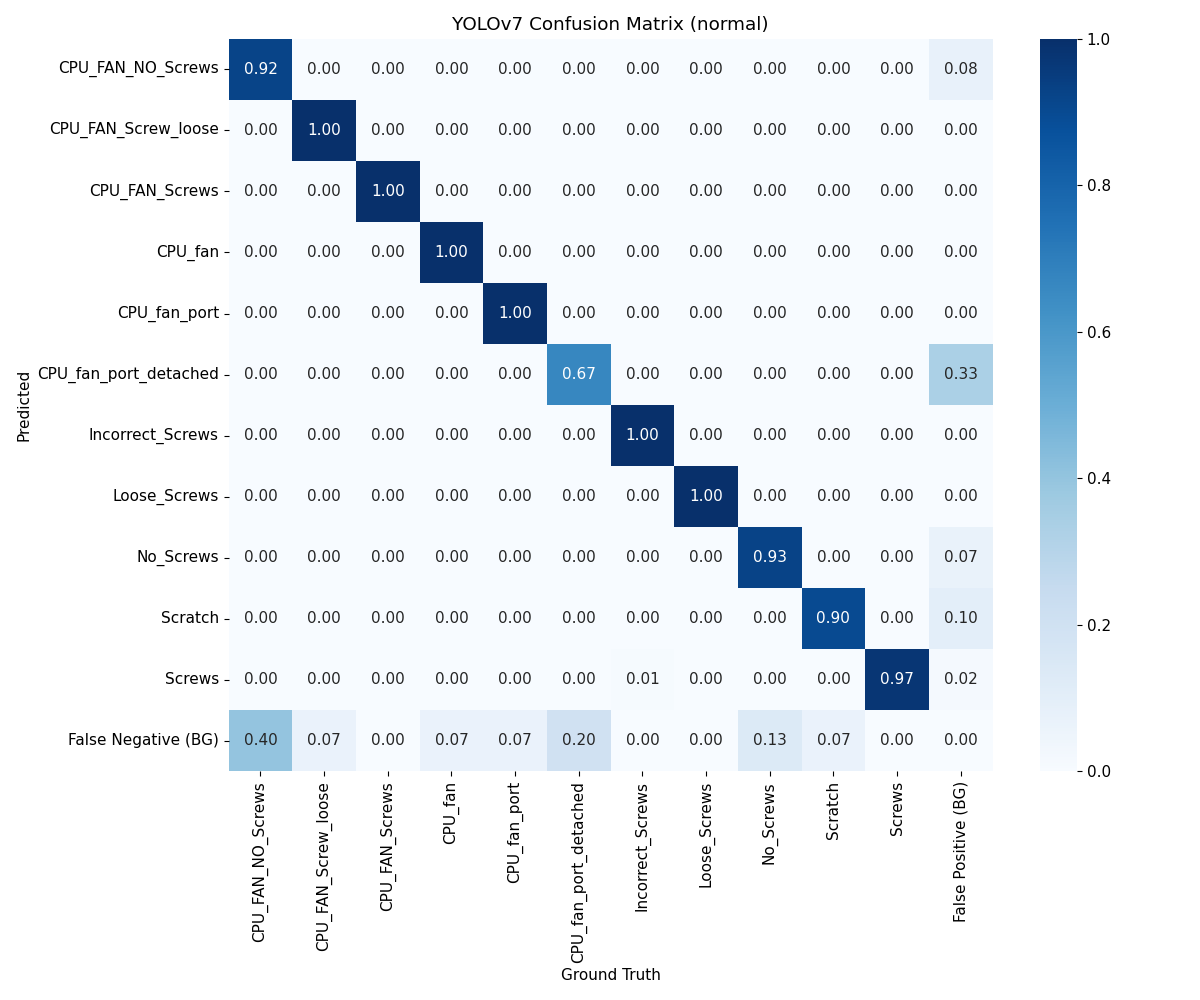}
    \caption{Normalized confusion matrix for YOLOv7 predictions.}
    \label{fig:yolov7_confusion_matrix}
\end{figure*}

\begin{figure*}[htbp]
    \centering
    \includegraphics[width=0.7\linewidth]{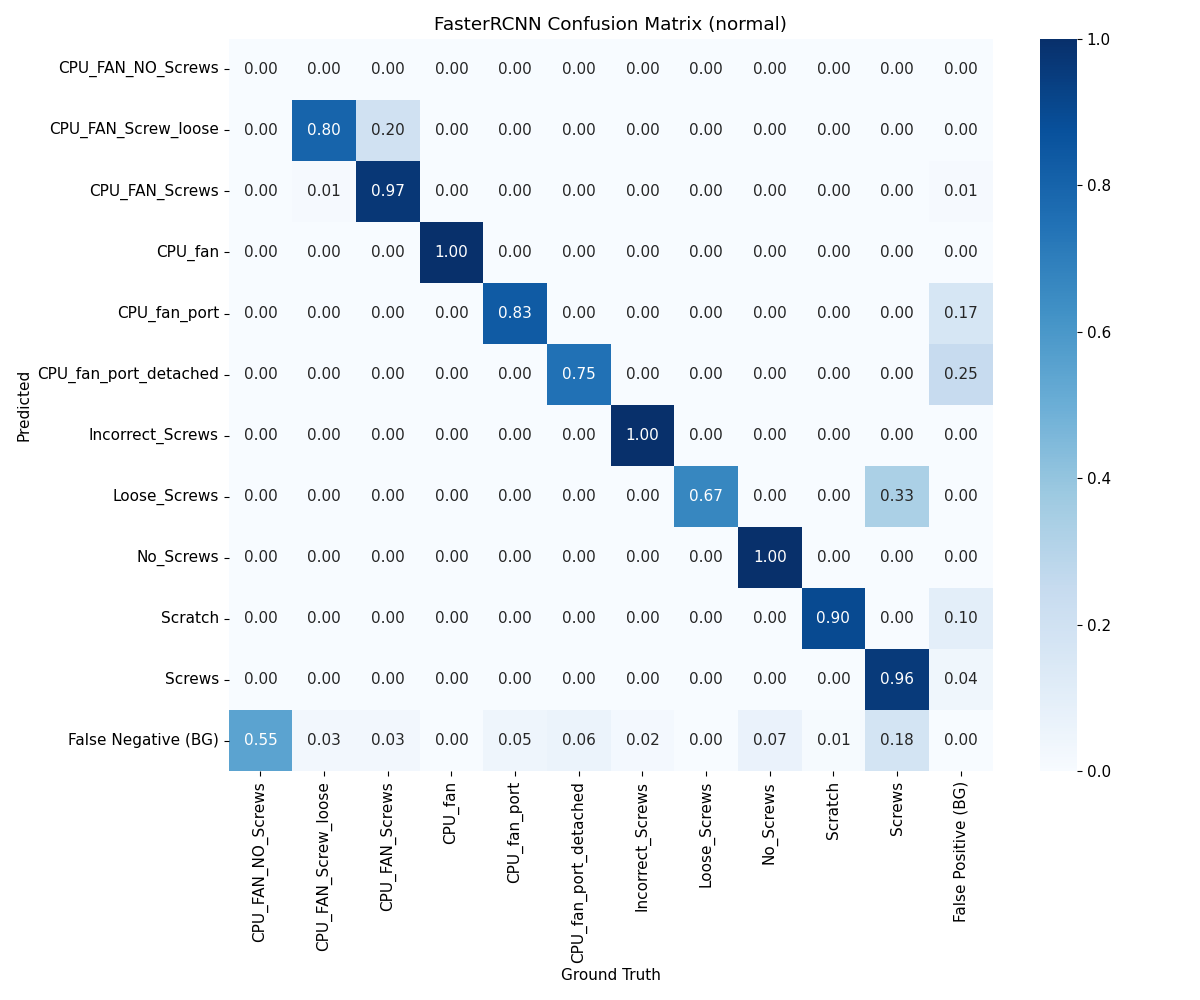}
    \caption{Confusion matrix for Faster R-CNN predictions.}
    \label{fig:fasterrcnn_confusion_matrix}
\end{figure*}

\begin{figure*}[htbp]
    \centering
    \includegraphics[width=0.7\linewidth]{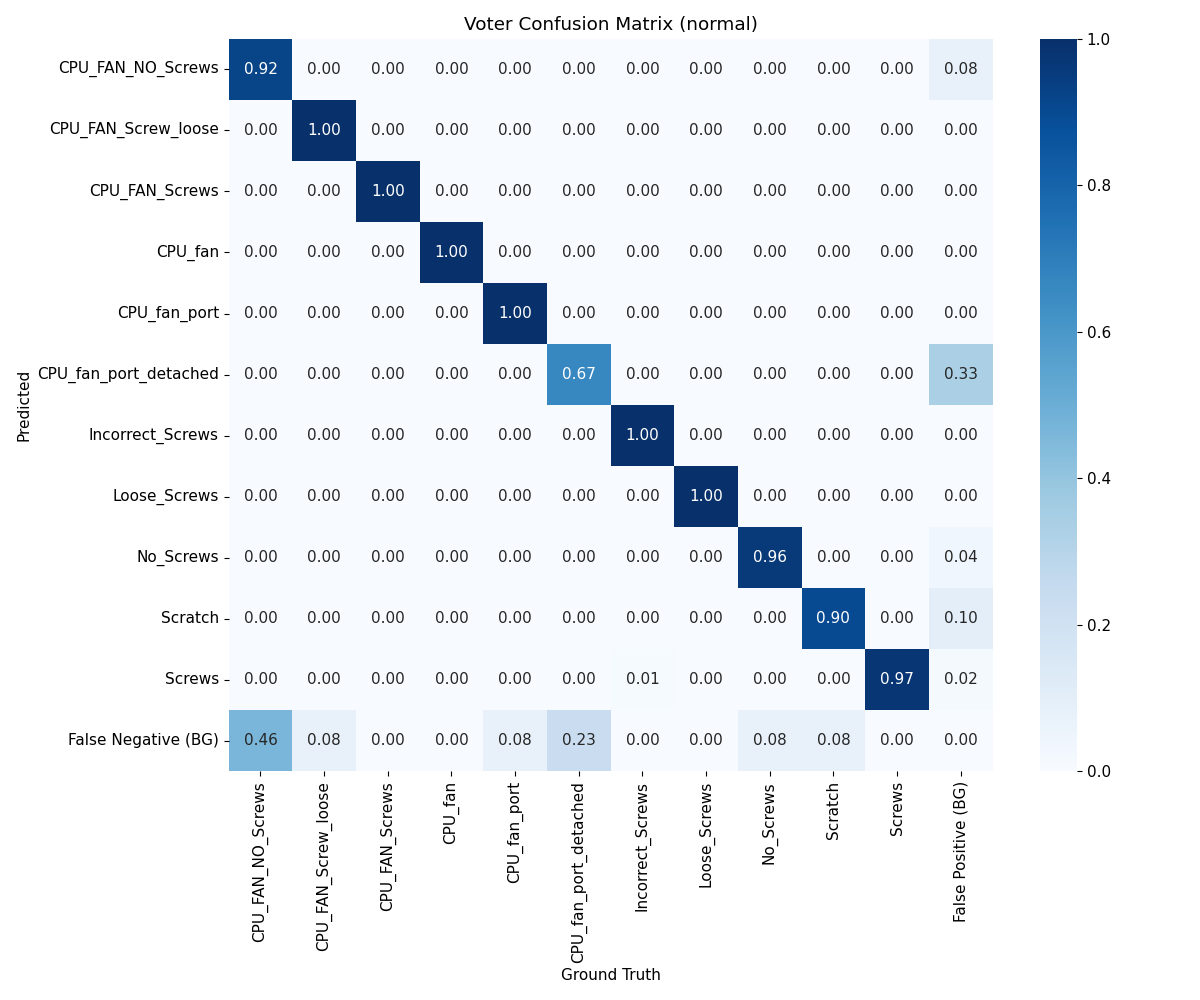}
    \caption{Confusion matrix for the proposed Voter Ensemble. Misclassifications are sparser, reflecting improved reliability on rare and ambiguous classes through ensemble fusion.}
    \label{fig:voter_confusion_matrix}
\end{figure*}

\begin{figure*}[htbp]
    \centering
    \includegraphics[width=0.95\linewidth]{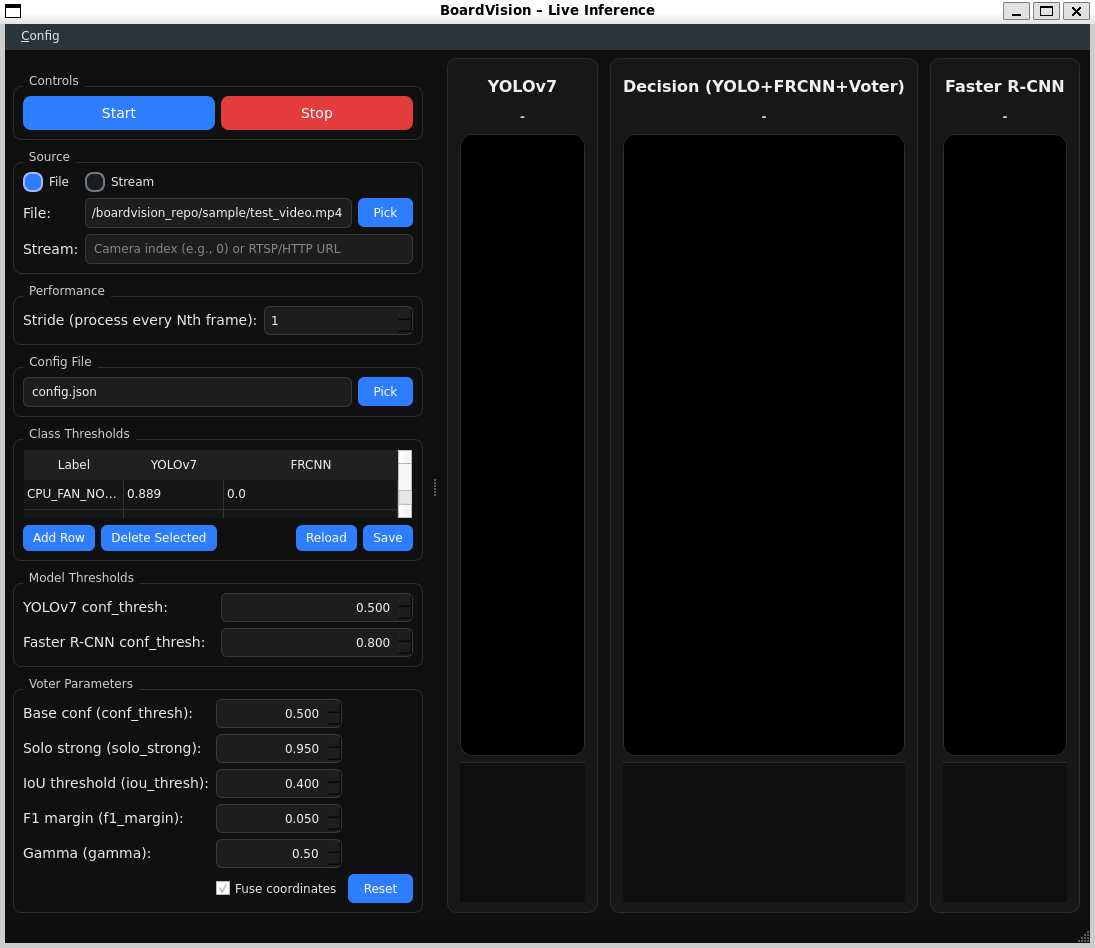}
    \caption{BoardVision GUI prior to selecting an input source, shown with synchronized YOLOv7, Faster R-CNN, and voter ensemble views.}
    \label{fig:boardvision_ui}
\end{figure*}

\end{document}